\pdfoutput=1

\documentclass[11pt]{article}
\usepackage[table]{xcolor}

\usepackage{acl}

\usepackage{times}
\usepackage[inline]{enumitem}
\usepackage{latexsym}
\usepackage{booktabs}
\usepackage{graphicx}
\usepackage{multirow}
\usepackage{makecell}
\usepackage{subcaption}
\usepackage{standalone}
\usepackage{pgfplots, pgfplotstable}
\pgfplotsset{compat=1.16}
\usepackage{float}

\usepackage[T1]{fontenc}

\usepackage[utf8]{inputenc}

\usepackage{microtype}

\usepackage{microtype}

\usepackage{xcolor}
\usepackage{amssymb}

\definecolor{mmcolor}{rgb}{0.1,0.2,0.4}
\definecolor{urjacolor}{rgb}{0.039, 0.478, 0.741}
\definecolor{selcolor}{rgb}{0.6, 0.2, 0.1}

\definecolor{todocolor}{rgb}{0.9,0.1,0.1}
\definecolor{changedcolor}{rgb}{0.42,0.27,0.57}
\definecolor{removedcolor}{rgb}{0.867,0.176,0.361}

\title{Will It Blend? Mixing Training Paradigms \& Prompting for Argument Quality Prediction}

\author{Michiel van der Meer \\ Leiden University \\ {\small \texttt{m.t.van.der.meer@liacs.leidenuniv.nl}} \\ \And  
  Myrthe Reuver\\
  Vrije Universiteit Amsterdam \\
  \texttt{myrthe.reuver@vu.nl} \\ \AND
  Urja Khurana \\
  Vrije Universiteit Amsterdam \\
  \texttt{u.khurana@vu.nl}\\ \And
  Lea Krause \\
  Vrije Universiteit Amsterdam \\
  \texttt{l.krause@vu.nl} \\ \And
  Selene Báez Santamaría \\
  Vrije Universiteit Amsterdam \\
  \texttt{s.baezsantamaria@vu.nl}
}

\begin{document}
\usetikzlibrary{patterns}
\maketitle
\begin{abstract}
This paper describes our contributions to the Shared Task of the 9th Workshop on Argument Mining (2022). Our approach uses Large Language Models for the task of Argument Quality Prediction. We perform prompt engineering using GPT-3, and also investigate the training paradigms multi-task learning, contrastive learning, and intermediate-task training. We find that a mixed prediction setup outperforms single models. Prompting GPT-3 works best for predicting argument validity, and argument novelty is best estimated by a model trained using all three training paradigms.
\end{abstract}

\section{Introduction}
As debates are moving increasingly online, automatically processing and moderating arguments becomes essential to further fruitful discussions. 
The research field of automatic extraction, analysis, and relation detection of argument units is called Argument Mining \citep[AM,][]{lawrence2020argument}. 

The shared task of the 9th Workshop on Argument Mining (2022) focuses on argument quality \citep{wachsmuth2017computational}. Argument quality can be broken down into multiple dimensions, each with its own purpose, or be extended to \emph{deliberative quality} \citep{vecchi2021towards}. The shared task includes two aspects of the \emph{logical} argument quality dimension: \emph{validity} and \emph{novelty}. Given a premise and a conclusion, a valid relationship indicates that sound logical inferences link the premise and conclusion. A novel relationship indicates that new information was introduced in the conclusion that was not present in the premise.
Prediction of an argument's validity and novelty can be either through binary classification (Task A) or by explicit comparison between two arguments (Task B). We focus on Task A. 

A system that is able to estimate validity and novelty could be a building block in AM for online deliberation. For instance, in assisting humans to detect arguments in online deliberative discussions \citep{vandermeer2022hyena, falk2021predicting} or presenting diverse viewpoints to users in a news recommendation system \citep{reuver2021we}.

We address the task of validity and novelty prediction through a variety of approaches ranging from prompting, contrastive learning, intermediate task training, and multi-task learning. Our best-performing approach is a mix of a GPT-3 model (through prompting) and a contrastively trained multi-task model that uses NLI as an intermediate training task. This approach achieves a combined Validity and Novelty F1-score of $0.45$.

\begin{figure*}[t]
    \centering
    \includegraphics[width=\textwidth]{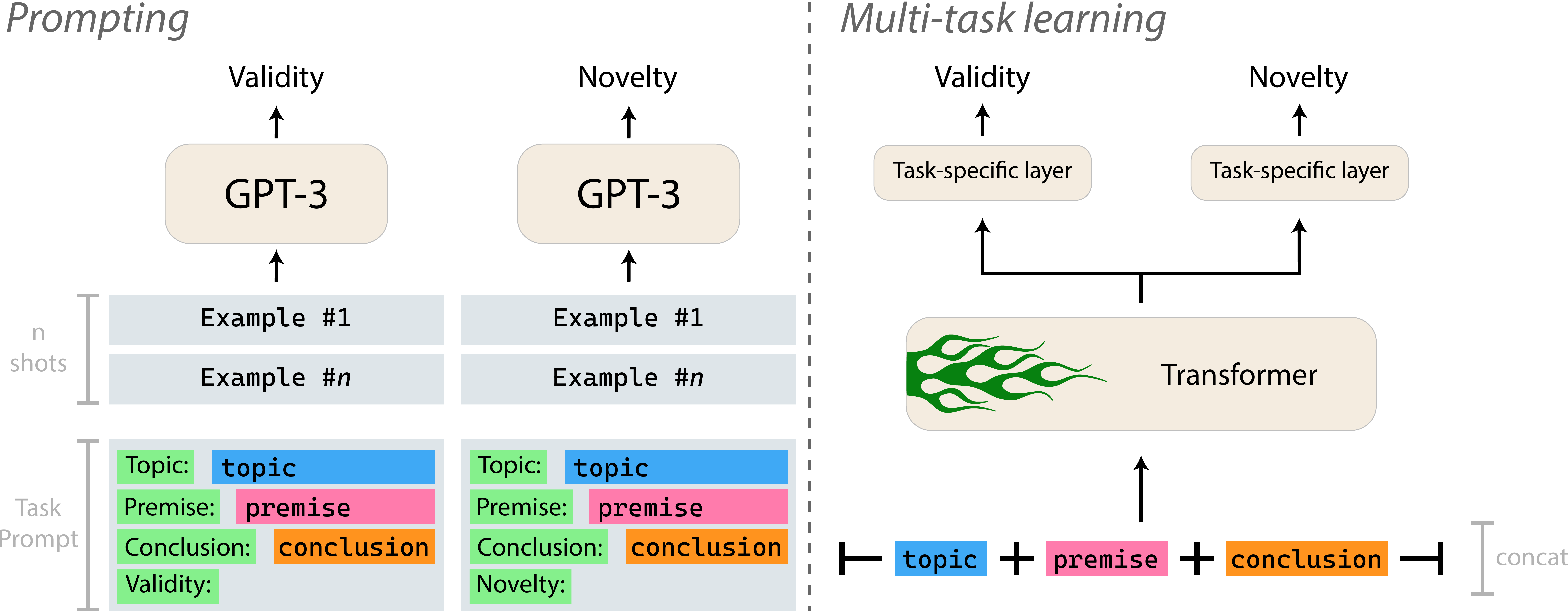}
    \caption{The two argument quality prediction setups used in our submissions. At inference time, predictions from different approaches may be mixed. }
    \label{fig:approaches}
\end{figure*}

\section{Related Work: Paradigms \& Prompting}
Given the two related argumentation tasks (novelty and validity), a Multi-Task Learning (MTL) setup \citep[]{crawshaw2020multi} is a natural approach. Multi-task models use training signals across several tasks, and have been applied before in argument-related work with Large Language Models (LLMs) \citep{lauscher2020rhetoric, cheng2020ape, tran2021multi}. We use shared encoders followed by task-specific classification heads. The training of these encoders was influenced by the following two lines of work. 

First, intermediate task training \cite{pruksachatkun-etal-2020-intermediate, weller2022use} fine-tunes a pre-trained LLM on an auxiliary task before moving on to the final task. This can aid classification performance, also in AM \citep{shnarch2022cluster}.

Second, contrastive learning is shown to be a promising approach \citep{alshomary2021key, phan2021matching} in a previous AM shared task \citep{friedman2021overview}. Contrastive learning is used to improve embeddings by forcing similar data points to be closer in space and dissimilar data points to be further away. Such an approach may cause the encoder to learn dataset-specific features that help in downstream task performance.

In addition to MTL, we look at prompt engineering for LLMs, which has shown remarkable progress in a large variety of tasks in combination with \citep{brown2020language} or without few-shot learning \citep{prompt-engineering}. For this task we draw inspiration from ProP \citep{selene-awesome-report-gpt}, an approach that ranked first in the ``Knowledge Base Construction from Pre-trained Language Models''  challenge at ISWC 2022.\footnote{LM-KBC, \url{https://lm-kbc.github.io/}} ProP reports the highest performance with 
\begin{enumerate*}[label=(\arabic*)]
    \item larger LLMs,
    \item shorter prompts,
    \item diverse and complete examples in the prompt,
    \item task-specific prompts.
\end{enumerate*}

\section{Data and Training Paradigms}

\subsection{Data}
The task data is in American English and consists of Premise, Conclusion, Topic, and a Novel and Validity label. As highlighted in Table~\ref{tab:data}, arguments that are both non-valid and novel are underrepresented in the data. We use the original training and validation distribution as provided and do not use any over- or undersampling strategies. Instead, we opt to resolve the data imbalance by adopting different training paradigms (see Section~\ref{sec:training}). 

\begin{table}[tb]
\small{
    \centering
    \begin{tabular}{@{}lrcrrrr@{}}
    \toprule
        \textbf{Split} & \textbf{Size} & \textbf{Distribution} & \textbf{Topics} & \multicolumn{2}{l}{\textbf{Topic Overlap}}\\
       &&&& \small{\textbf{w. train}} & \small{\textbf{w. dev}}\\
    \midrule
         train  & 750 & 331/{\color{red} 18}/296/105 & 22 &  -- & 0\\
         dev    & 202 & 33/44/87/38 & 8 &  0 & --                    \\
         test   & 520 & 110/96/184/130 & 15 &  0 & 8              \\
    \bottomrule
    \end{tabular}
    }
    \caption{Shared task data overview. \textbf{Distribution} indicates the class distribution of $\{$non-valid, non-novel$\}$/$\{$non-valid, novel$\}$/$\{$valid, non-novel$\}$/$\{$valid,novel$\}$ counts. The red count indicates a severe data imbalance in the training set.}
    \label{tab:data}
\end{table}

The content included in the dataset concerns common controversial issues popular on debate portals \citep{gretz2020large}, with topics varying from ``TV Viewing is Harmful to Children'' to "Turkey EU Membership''.

The training data also contains classes labelled ``defeasibly'' valid and ``somewhat'' novel, which are not in the development or test set. We map these to negative labels (i.e. not novel or not valid) to refrain from discarding data. However, we do not measure the effect of this decision on performance.

\subsection{Training Paradigms}
\label{sec:training}
In our submissions, we mix different training paradigms to obtain our final approach. A schematic overview is given in Figure~\ref{fig:approaches}. Below, we outline each of the paradigms individually. 

\paragraph{Multi-task Learning}
Since both validity and novelty are related, a shared encoder is used to process the text input into an embedding, which is fed to task-specific layers. We do not use any parameter freezing, allowing gradients from either task to pass through the entire encoder. During training, a single task is sampled uniformly at random, and a batch is sampled containing instances for that task.

\paragraph{Intermediate task training} 
In our case, we use two related tasks for intermediate task training: Natural Language Inference (NLI) and argument relation prediction. For NLI, we use a released RoBERTa model \cite{liu2019roberta} trained on the MNLI corpus \citep{williams2018broad}, predicting whether two sentences show logical entailment. This is related because making sound logical inferences plays a role in validity. The released argument relation RoBERTa model \citep{ruiz2021transformer} was trained on the relationship (inference, contradiction, or unrelated) between two sentences in a debate \citep{visser2020argumentation}. This is related to novelty and validity. For instance, unrelated arguments may be novel but not valid, and vice versa.

\paragraph{Contrastive Learning}
We use SimCSE's \citep{gao2021simcse} supervised setting to further fine-tune the previously mentioned RoBERTa MNLI model in a contrastive manner. To train the model we take triples of premises and conclusions in the form of premise, conclusion with a positive novelty rating, and conclusion with a negative novelty rating.

\section{Approach}
\label{sec:approach}
\subsection{Submitted Approaches}\label{sec:submit}

\paragraph{Approach 1: GPT-3 Prompting}
In our prompt-engineering approach, we use OpenAI's GPT-3\footnote{\url{https://beta.openai.com/playground}} \cite{brown2020language} for few-shot classification of novelty and validity labels. We construct a prompt by concatenating the topic, premise, and conclusion in a structured format, and request either a validity or novelty label in separate prompts. In addition, we show four static examples before asking for a label from the model, selected from short, difficult examples (i.e. those with the lowest annotation agreement) in the training dataset.
    
\paragraph{Approach 2: NLI as Intermediate-task, Contrastive learning and Multi-Task Learning}
This model consists of a shared encoder with task-specific classification heads. We initialize the shared encoder using a pretrained RoBERTa model on the MNLI corpus. We then perform contrastive learning with a triplet loss. Afterward, the model is fine-tuned using MTL on the shared task training data. During training, we switch uniformly at random during training between the novelty and validity tasks.

\paragraph{Approach 3: Mixing Approach 1 (GPT-3) \& Approach 2 (NLI+contrastive+MTL)}
Our Mixed Approach uses Approach 1 (prompt engineering) for validity labels, and Approach 2 (fine-tuned model) for novelty labels.
    
\paragraph{Approach 4: ArgRel as Intermediate-task and Multi-Task Learning}
This model uses intermediate-task training on the argument relation prediction task followed by Multi-Task Learning in the same set-up as in Approach 1, but without contrastive learning.

\paragraph{Approach 5: Mixing Approach 1 (GPT-3) \& Approach 4 (ArgRel+MTL)}
This approach uses Approach 1 (prompt engineering) for validity and Approach 4 (ArgRel+MTL) for novelty labels.

\begin{table*}[h!]
    \centering
    \begin{tabular}{@{}lccc@{}}
        \toprule
        \multirow{2}{*}{\textbf{Model}} & \multicolumn{3}{c}{\textbf{F1}}\\
        \cmidrule(ll){2-4}
         & \textbf{Validity} & \textbf{Novelty} & \textbf{Combined}\\
        \midrule
        \textbf{SVM} (TF-IDF + stemming)            & 0.60 & 0.08    & 0.21 \\
        \textbf{GPT-3} (CLTeamL-1)                  & 0.75 & 0.46 & 0.35  \\
        \textbf{NLI+contrastive+MTL} (CLTeamL-2)    & 0.65 & 0.62 & 0.39  \\
        \textbf{GPT-3 \& NLI+contrastive+MTL} (CLTeamL-3)*    & \textbf{0.75} & \textbf{0.62} & \textbf{0.45}  \\
        \textbf{ArgRel+MTL} (CLTeamL-4)             & 0.57 & 0.59 & 0.33  \\
        \textbf{GPT-3 \& ArgRel+MTL} (CLTeamL-5)      & 0.75 & 0.59 & 0.43  \\
        \bottomrule
    \end{tabular}
    \caption{Test set performance. CLTeamL-\textit{n} indicates an official submission with \textit{n} corresponding to the Approach number also in Section~\ref{sec:submit}. Bold scores indicate the best-performing approach in the shared task. "Combined" indicates the Shared Task organizer's scoring metric for both tasks.}
    \label{tab:scores-test}
\end{table*}

\subsection{Non-submitted Approaches}
\paragraph{Baseline: SVM}
Support Vector Machines (SVMs) are strong baselines for argument mining tasks with relatively small multi-topic datasets \cite{reuver2021stance}. We train an SVM separately for validity and novelty as a competitive baseline.

\subsection{Implementation details}
We use Python3 and the HuggingFace \texttt{transformers} \citep{wolf2020transformers} framework for training our models. The SVM baseline instead uses sklearn \citep{scikit-learn}. Our code is publicly available.\footnote{\url{https://github.com/m0re4u/argmining2022}} All models trained use RoBERTa (large) \citep{liu2019roberta} as the base model, and the intermediate task trained models are obtained directly from the HuggingFace Hub.\footnote{\url{https://huggingface.co/}} We provide hyperparameters for fine-tuned trained models in Appendix~\ref{app:hyperparameters}.

Model selection was done based on the combined (validity and novelty) F1 performance on the development set. All experiments were run for 10 epochs, after which the best-performing checkpoint was selected for use in creating predictions on the test set. The training was performed on machines including either two GTX2080 Ti GPUs, or four GTX3090 GPUs.

\section{Experiments and Results}

We compare our approaches' performance on the test set with the shared task's metric (Combined F1 of Validity and Novelty). 
Additionally, we analyze our approaches' errors and their connection to labels, annotator confidence, and topic. 

\subsection{Test set performance}\label{Test}
See Table~\ref{tab:scores-test} for performance on the test set. We also present a not-submitted SVM as a baseline.

\subsection{Error Analysis}
We perform additional error analysis on three approaches (Approach 1, 2, and 3). We analyze errors in terms of \begin{enumerate*}[label=(\arabic*)]
\item label-specific performance,
\item annotator confidence, and
\item topics. 
\end{enumerate*} Additional results are in Appendix~\ref{app:addlt}.

\paragraph{Per-label performance}\label{class} We observe complementary strengths for the GPT-3 model and our MTL approach in Tables~\ref{tab:results-per-label}. The MTL model is remarkably stronger than GPT-3 at identifying \emph{novel} arguments, even when considering this is a low-frequency class. We see a similar trend in terms of misclassifications (Table~\ref{tab:confus-nov}), as the MTL model has a 40\% lower error rate for the novelty label. 

\begin{table}[H]
    \centering
    \begin{tabular}{@{}lcccc@{}}
        \toprule
        \multirow{2}{*}{\textbf{Model}} & \multicolumn{2}{c}{\textbf{F1 Validity}} & \multicolumn{2}{c}{\textbf{F1 Novelty}}\\
        \cmidrule(ll){2-3} \cmidrule(ll){4-5}
         & valid & non-valid & novel& non-novel\\
        \midrule
        \textbf{GPT-3}   & 0.78 &  0.62 & 0.28 & 0.67 \\
        \textbf{MTL}     & 0.80 &  0.50 & 0.48 & 0.75 \\
        \bottomrule
    \end{tabular}
    \caption{Per-label performance on the test set. }
    \label{tab:results-per-label}
\end{table}

\begin{table}[H]
    \begin{minipage}{.5\linewidth}
      \centering
    \begin{tabular}{@{}clcc@{}}
         \toprule
         && \multicolumn{2}{c}{Predicted}\\
         && \textbf{-} & \textbf{+}\\
        \midrule
         \parbox[t]{2mm}{\multirow{2}{*}{\rotatebox[origin=c]{90}{True}}}&\textbf{-}  & 240 & 54\\
         &\textbf{+}  & 181 & 45\\
        \bottomrule
    \end{tabular}
        \subcaption{GPT-3}

    \end{minipage}%
    \begin{minipage}{.5\linewidth}
      \centering
    \begin{tabular}{@{}clcc@{}}
         \toprule
         && \multicolumn{2}{c}{Predicted}\\
         && \textbf{-} & \textbf{+}\\
        \midrule
         \parbox[t]{2mm}{\multirow{2}{*}{\rotatebox[origin=c]{90}{True}}}&\textbf{-}  & 265 & 29\\
         &\textbf{+}  &145 &81\\
        \bottomrule
    \end{tabular}
        \subcaption{MTL}
    \end{minipage} 
    \caption{Confusion matrices for the novelty labels.}
    \label{tab:confus-nov}
\end{table}

\begin{figure}[t]
    \centering
    \quad\quad\ref{namedlegend}
    \begin{minipage}{.5\linewidth}
  \centering
\begin{tikzpicture}[font=\sffamily,scale=0.45]
    \begin{axis}[
            ylabel={Relative Freq.},
            grid,
            ybar,
            bar width=10pt,
            legend columns=2,
            legend entries={mistakes (GPT-3), mistakes (MTL), correct (GPT-3), correct (MTL)},
            legend cell align={left},
            legend to name=namedlegend,
            enlarge x limits={abs=30pt},
            xtick={0,1,2},
            xticklabels={Very confident, majority, confident}
        ]
    \addplot[
        red!80,fill=red!20
    ] table [x index=0, y index=1] {images/data_conf_val1.dat};
    \addplot[
        red!80,fill=red!20,
        pattern=north east lines,
        pattern color=red!60
    ] table [x index=0, y index=1] {images/data_conf_val2.dat};
    \addplot[
        blue!80,fill=blue!20
    ] table [x index=0, y index=2] {images/data_conf_val1.dat};
    \addplot[
        blue!80,fill=blue!20,
        pattern=north east lines,
        pattern color=blue!60,
    ] table [x index=0, y index=2] {images/data_conf_val2.dat};
    \end{axis}

\end{tikzpicture}
\subcaption{Validity}
    \end{minipage}%
    \begin{minipage}{.5\linewidth}
        \centering
\begin{tikzpicture}[font=\sffamily,scale=0.45]
    \begin{axis}[
            ylabel={Relative Freq.},
            grid,
            ybar,
            bar width=10pt,
            enlarge x limits={abs=30pt},
            xtick={0,1,2},
            xticklabels={Very confident, majority, confident}
        ]
    \addplot[
        red!80,fill=red!20
    ] table [x index=0, y index=1] {images/data_conf_nov1.dat};
    \addplot[
        red!80,fill=red!20,
        pattern=north east lines,
        pattern color=red!60
    ] table [x index=0, y index=1] {images/data_conf_nov2.dat};
    \addplot[
        blue!80,fill=blue!20
    ] table [x index=0, y index=2] {images/data_conf_nov1.dat};
    \addplot[
        blue!80,fill=blue!20,
        pattern=north east lines,
        pattern color=blue!60,
    ] table [x index=0, y index=2] {images/data_conf_nov2.dat};
    \end{axis}

\end{tikzpicture}
\subcaption{Novelty}
    \end{minipage}

    \caption{Relative accuracy rates divided over label confidence scores.}
    \label{fig:lconf-1}
\end{figure}
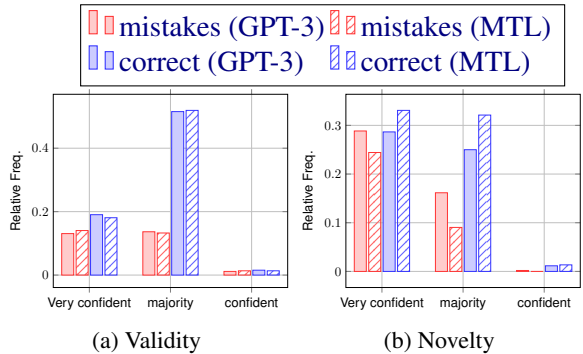

\paragraph{Annotator confidence}\label{annotator} 
See Figure~\ref{fig:lconf-1} for the relationship between annotator confidence and classification error. Surprisingly, examples labeled as very confident (easy for human annotators) are not consistently correctly classified by any approach. For novelty, GPT-3 gets about half of these examples wrong. 

\paragraph{Topics}\label{topic} 
The 3 topics with the highest error rates 
differ between approaches and tasks. For validity, GPT-3 struggles with ``Was the Iraq War Worth it?" (44.8\%), while MTL with ``Vegetarianism" (40\%). For novelty, GPT-3 also struggles with
"Vegetarianism" (60\%), and MTL with ``Withdrawing from Iraq" (44.7\%) and ``Vegetarianism" (44\%).

\section{Conclusion}
We highlight two main conclusions.

(1) \textbf{Different models have different strengths relating to the two tasks}. A prompting approach with a generative model worked best for validity, while contrastive supervised learning worked best for novelty. The two tasks are related enough to be able to effectively use one multi-task learning model, but merging predictions from multiple heterogeneous models leads to the best score.

(2) \textbf{Specific intermediate-tasks before fine-tuning work well for low-resource argument mining tasks}. NLI seems clearly related to validity prediction. For the novelty tasks, other tasks related to argument similarity \citep{reimers2019classification} might be equally informative.

\section{Access and Responsible Research}
A core consideration in NLP research when sharing results is the accessibility and reproducibility of the solution. While our code is openly available, the approaches including GPT-3 require access to commercially trained models. We used free trial OpenAI accounts (allowing \$18 of free GPT-3 credit), but larger datasets and additional tasks can quickly make this approach infeasible. We also considered the freely accessible LLM BLOOM\footnote{\url{https://huggingface.co/bigscience/bloom}}. BLOOM does not require payment, but does require more GPU memory than what was available to us -- making it inaccessible. 

Ultimately, GPT-3 and related LLMs have several biases and risks of use, including the generation of false information \cite{tamkin2021understanding} and the fact that their training on internet language leads to a very limited set of language, ideas, and perspectives represented \cite{bender2021dangers}, with even racist, sexist, and hateful views \cite{gehman2020realtoxicityprompts}. This is especially important to mention, as the task description mentions a future use case of generating new arguments.

\newpage
\section*{Acknowledgements}
This research was funded by the Vrije Universiteit Amsterdam and the Netherlands Organisation for Scientific Research (NWO) through the \textit{Hybrid Intelligence Centre} via the Zwaartekracht grant (024.004.022), the \textit{Rethinking News Algorithms} project via the Open Competition Digitalization Humanities \& Social Science grant (406.D1.19.073), and the \textit{Spinoza} grant (SPI 63-260) awarded to Piek Vossen. We would also like to thank the reviewers for their excellent feedback that enhanced this paper. All remaining errors are our own.  

\bibliography{custom}

\newpage
~\newpage

\appendix
\section{Hyperparameters}
\label{app:hyperparameters}
\paragraph{GPT-3 Prompt}
We used the model \texttt{text-davinci-002} with a temperature of 0 and no penalties on frequency and presence. We experimented with various prompt designs (e.g. dynamic or longer examples, more/fewer examples, joint prompting of novelty and validity) but manual inspection showed the best results for the present setup described in the paper (i.e. separate prompts, static prompt style). 

\paragraph{Transformers}
We report the hyperparameters for each approach in Table~\ref{tab:hyper} that differ from the default. In all Transformer models, we used the AdamW optimizer \citep{loshchilov2018decoupled}.
\begin{table}[!htb]
    \centering
    \begin{tabular}{@{}lccc@{}}
        \toprule
        \textbf{Model} & \textbf{LR} & \textbf{epochs} & \textbf{g.acc.} \\
        \midrule
         CLTeamL-2 & 1e-05 & 9 & 1\\
         CLTeamL-3 (novelty) & 1e-05 & 9 & 1\\
         CLTeamL-4 & 5e-06 & 6 & 4\\
         CLTeamL-5 (novelty) & 5e-06 & 6 & 4\\
        \bottomrule
    \end{tabular}
    \caption{Hyperparameters for our approaches that involve gradient-based learning.}
    \label{tab:hyper}
\end{table}

\paragraph{SVM}
The best performing model on the validation set is one with a C parameter of 0.09 for validity and 4.7 for novelty. 
The text representation concatenates the two texts, in a TF-IDF and stemmed (with the SnowBall stemmer as implemented in NLTK) representation.

\section{Additional results}
For every analysis, we show the results for approaches \emph{CLTeamL-1} and \emph{CLTeamL-2}, which can be combined into \emph{CLTeamL-3} by merging their results (take validity and novelty, respectively for \emph{1} and \emph{2}).
\label{app:addlt}
\subsection{Per-label Performance}
See Tables~\ref{tab:stats-best-model-1-per-label} and~\ref{tab:stats-best-model-2-per-label}.
\begin{table}[tb]
    \centering
    \begin{tabular}{@{}lcccc@{}}
         \toprule
         & \textbf{Prec.}& \textbf{Rec.} & \textbf{F1} & \textbf{Support}  \\
        \midrule
         non-valid   & 0.583 & 0.670 & 0.623 & 179 \\
         valid       & 0.812 & 0.748 & 0.779 & 341\\
        \midrule
         non-novel   & 0.816 & 0.570 & 0.671 & 421\\
         novel       & 0.199 & 0.455 & 0.277 & 99\\
        \bottomrule
    \end{tabular}
    \caption{Performance statistics for approach \emph{CLTeamL-1}.}
    \label{tab:stats-best-model-1-per-label}
\end{table}

\begin{table}[tb]
    \centering
    \begin{tabular}{@{}lcccc@{}}
         \toprule
         & \textbf{Prec.}& \textbf{Rec.} & \textbf{F1} & \textbf{Support}  \\
        \midrule
         non-valid   & 0.364 & 0.806 & 0.502 & 93 \\
         valid       & 0.943 & 0.693 & 0.799 & 427\\
        \midrule
         non-novel   & 0.901 & 0.646 & 0.753 & 410\\
         novel       & 0.358 & 0.736 & 0.482 & 110\\
        \bottomrule
    \end{tabular}
    \caption{Performance statistics for approach \emph{CLTeamL-2}.}
    \label{tab:stats-best-model-2-per-label}
\end{table}

\subsection{Label confusion} 
See Tables~\ref{tab:confus-nov} and~\ref{tab:confus-1}.

\begin{table}[tb]
    \begin{minipage}{.5\linewidth}
      \centering
    \begin{tabular}{@{}clcc@{}}
         \toprule
         && \multicolumn{2}{c}{Predicted}\\
         && \textbf{-} & \textbf{+}\\
        \midrule
         \parbox[t]{2mm}{\multirow{2}{*}{\rotatebox[origin=c]{90}{True}}}&\textbf{-}  & 120& 86\\
         &\textbf{+}  & 59 & 255 \\
        \bottomrule
    \end{tabular}
      \subcaption{GPT-3}

    \end{minipage}%
    \begin{minipage}{.5\linewidth}
      \centering
    \begin{tabular}{@{}clcc@{}}
         \toprule
         && \multicolumn{2}{c}{Predicted}\\
         && \textbf{-} & \textbf{+}\\
        \midrule
         \parbox[t]{2mm}{\multirow{2}{*}{\rotatebox[origin=c]{90}{True}}}&\textbf{-}  & 75& 131\\
         &\textbf{+}  & 18 & 296 \\
        \bottomrule
    \end{tabular}
      \subcaption{MTL}
    \end{minipage} 
    \caption{Confusion matrices for the validity labels.}
    \label{tab:confus-1}
\end{table}

\subsection{Seed Variance}
\begin{figure}[h!]
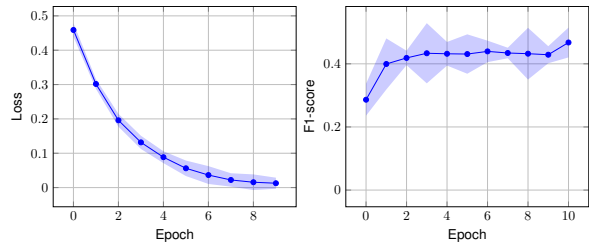

    \begin{minipage}{.5\linewidth}
  \centering
    \includestandalone[mode=tex, width=\columnwidth]{images/seeds_loss}
    \end{minipage}%
    \begin{minipage}{.5\linewidth}
        \centering
    \includestandalone[mode=tex, width=\columnwidth]{images/seeds_f1}
    \end{minipage} 
    \caption{Training loss and combined F1 score for multiple training runs of \emph{CLTeamL-2} with different seeds.}
    \label{fig:seeds}
\end{figure}

While the results for the task were obtained using a single model, we investigate training stability over multiple seeds. We show the results and variance from five different seeds for our best-performing MTL model. The results can be seen in Figure~\ref{fig:seeds}. Training is relatively stable, but individual models may have small performance differences on the test set.

 \subsection{Topics}\label{appendix:topics}
The three most error-prone topics were different for approaches. Notable is that ``Vegetarianism" is an error-prone topic across tasks and approaches.

\paragraph{GPT-3 - Validity} ``Was the Iraq War Worth it?" (unseen) with 44.8\% errors, ``Year Round School" (unseen), 39.7\% errors, and ``Withdrawing from Iraq" (unseen), 38.1\% errors.

\paragraph{GPT-3 - Novelty}
``Yucca Mountain nuclear waste" (62.5\% error rate), 
``Vegetarianism" (60\% error rate), ``Wiretapping in the U.S. (59.2\% error rate).

\paragraph{MTL - Validity}
``Zero Tolerance Law" (42.1\%), ``Vegetarianism" (40\% error rate) and ``Yucca Mountain nuclear waste" (37.5\% error rate).

\paragraph{MTL - Novelty}
``Withdrawing from Iraq" (44.7\% error rate), ``Vegetarianism" (44\% error rate), ``Wiretapping in the United States" (44\% error rate) 

\paragraph{Topics not in dev, only in test} ``Video games', ``Zero tolerance law', ``Was the War in Iraq worth it?', ``Withdrawing from Iraq', ``Year-round school', ``Veal', ``Water privatization'.

\end{document}